# Deep Learning Based Vehicle Tracking System Using License Plate Detection And Recognition

Lalit Lakshmanan[1], Yash Vora[2], Raj Ghate[3], Sarika Rane[4], Rajshekar Punna[5]

[123] *Student, Computer Engineering Department, Shah and Anchor Kutchhi Engineering College, Mumbai,Maharashtra,India*

[4] *Assistant Professor, Computer Engineering Department, Shah and Anchor Kutchhi Engineering College,Mumbai,Maharashtra,India*

[5] *Scientific Officer/D, Bhabha Atomic Research Centre, Mumbai, Maharashtra, India*

*Abstract*—Vehicle tracking is an integral part of intelligent traffic management systems. Previous implementations of vehicle tracking used Global Positioning System(GPS) based systems that gave location of an individual's vehicle on their smartphones. The proposed system uses a novel approach to vehicle tracking using Vehicle License plate detection and recognition (VLPR) technique, which can be integrated on a large scale with traffic management systems. Initial methods of implementing VLPR used simple image processing techniques which were quite experimental and heuristic. With the onset of Deep learning and Computer Vision, one can create robust VLPR systems that can produce results close to human efficiency. Previous implementations, based on deep learning, made use of object detection and support vector machines for detection and a heuristic image processing based approach for recognition. The proposed system makes use of scene text detection model architecture for License plate detection and for recognition it uses the Optical character recognition engine (OCR) Tesseract. The proposed system obtained extraordinary results when it was tested on a highway video using NVIDIA GeForce RTX 2080ti GPU, results were obtained at a speed of 30 frames per second with accuracy close to human.

## I. INTRODUCTION

With the rise of the automobile industry vehicles have been a common sight on the road and tracking them is one such security measure required to be taken. Vehicle Tracking is a very tedious and mind numbing task for the traffic management officers. Even with the use of cameras this task is difficult for a set of humans. Simple tasks such as finding a stolen vehicle by going through hours of footage can become mountainous if performed by humans alone. A vehicle tracking system can help not just in completing simple tasks but also complex tasks such as finding and charging fines against rule breakers, identifying traffic violators, finding illegal vehicles etc. Vehicles are identified by their license plate which is unique for every vehicle, hence tracking vehicles using a license plate is an efficient method.

The proposed system uses scene text detection technique for localising or detecting license plate and OCR engine for recognising the license plate details. These details are then stored in a database along with timestamp and at the location where it was spotted. Using the timestamp and the location a rough path traveled by the vehicle can be mapped out.

The applications and scope of the system is endless, example by integrating it with the RTO database one can get details of who a car belongs to and other essential details.

## II. EXISTING WORKS

Vehicle license plate detection and recognition problems have seen many solutions using different domains of computer science. Some systems use heuristic image processing methods, while others rely on deep neural networks and machine learning models.

An image processing based approach is proposed by Prof. Pradnya Randive et al, where for license plate detection purpose the concept of edge detection, contour determination and bounding box formation and elimination is used. Selection of license plate areas (LPA) and their elimination to obtain the actual license plate based on various heuristics is performed to filter out the improper detection. For the purpose of character recognition a properly segmented license plate is taken as input and some preprocessing is done on the license plate image for the removal of noise The noise free output image is sent for character segmentation. Further, Character recognition is done using template matching and license plate are authenticated using rank based search strategy[1]

A robust real time VLPR system is proposed by Rayson Laroca et al. In this system they have used You Only Look Once (YOLO) object detection architecture for localising license plates followed by character recognition and segmentation using Convolutional Neural Networks(CNN). In the SSIG dataset, composed of 2,000 frames from 101 vehicle videos, the system achieved a recognition rate of 93.53% and 47 Frames Per Second (FPS), performing better than both Sighthound and OpenALPR.[2]

Other methods for detecting license plates include using connected component analysis[3][4][5], 2D cross correlation [6] and Vector quantisation[7] which are all based on text detection.

All the above mentioned methods have their pros and cons. The first method that is mentioned is easy to implement but not very efficient as it is based on heuristic assumptions. Next method is YOLO based, which is quite efficient and fast. The data required to train an object detector module is huge and therefore it is not easy to implement.Also, there is difficulty faced by object detection in locating license plates that are painted on the vehicles, for example, auto-rickshaws, buses, etc. Therefore, the proposed system suggests a text detection based method for localisation of license plate as well as OCR for text recognition.

## III. SYSTEM OVERVIEW

The proposed system has a website for its frontend purpose, when a user opens the website he's asked to log in to the system using the username and password provided to it. Also every user is allocated roles - admin,basic- only two roles for now. An admin user is tasked with adding new users, adding new cameras, deleting the same. The basic user profile can view the regularly updated database and also view the live stream.

On the backend there are two models running on every node i.e. a camera location. The first model takes frames from the camera, and tries to locate the license plate. When the system gets a located license plate it sends the frame to the character recognition module where some preprocessing is done then the output is sent to the tesseract engine, which gives the characters of the license plate. The tesseract engine is trained using license plates first to improve its accuracy.

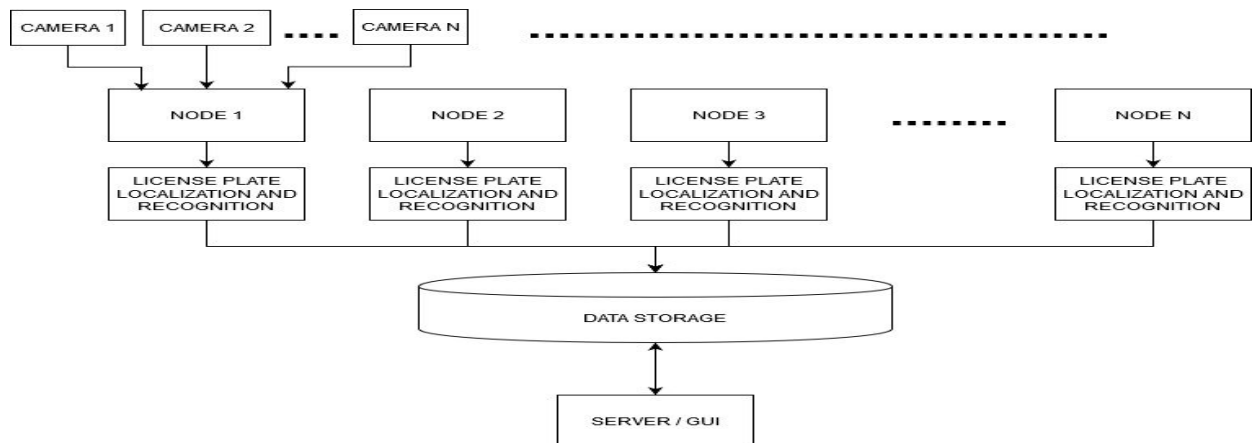

Fig 1. System Overview

## IV. SYSTEM REQUIREMENT

The system's frontend is built using Django Framework and the models are built using Keras machine learning Framework.

In order to host the website on a server, a Python development environment is needed along with necessary packages. Atleast, Python 3.6 version or higher is required along with packages like Numpy, OpenCV, Django, Pillow, ArgParse, Keras, Tensorflow.

For the database the system has used the PostGreSql database, also its Python client psycopg2 installed on the hosting server.

For users to access the service only need an internet connection and a browser

## V. METHODOLOGY

The proposed system is divided into three phases, first phase is the GUI and database i.e. the website and the second phase is license plate detection and the third phase is license plate recognition module.

### A. First Phase: GUI and Database

Web based GUI is used in the system as it can be accessed on any platform and device. The System uses the Django framework which is based on Python for the front-end, as it helps in creating dynamic websites. All other modules of the proposed system are in Python, so no integration overhead.

Another advantage of using Django is its admin panel which can be used to update the database, like adding users, deleting users etc.

For databases the system uses PostgreSql database, it is a free database service, it comes with a GUI panel for managing databases and is easy to integrate with Django.

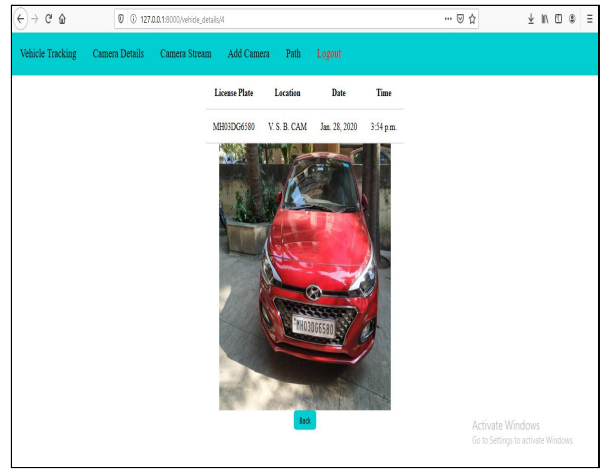

Fig 3a. Snapshots of GUI(1)

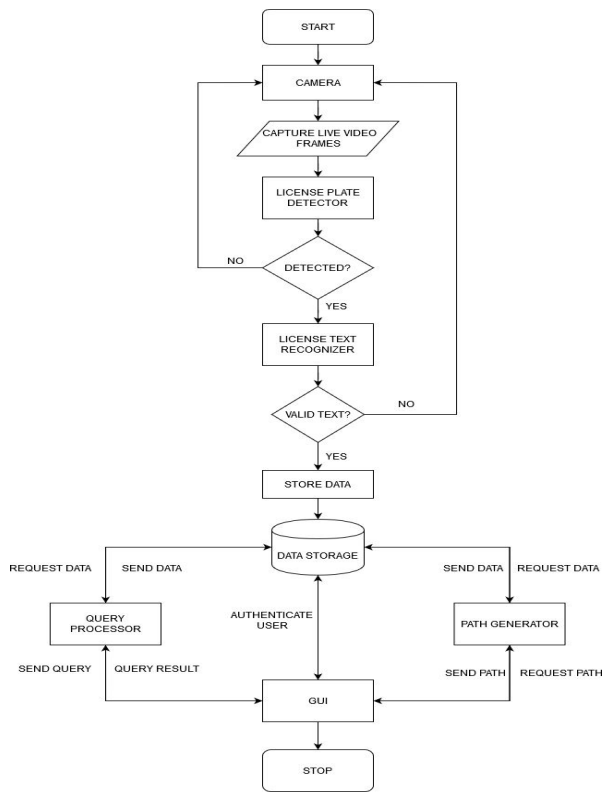

Fig 2. Flowchart of the Proposed System

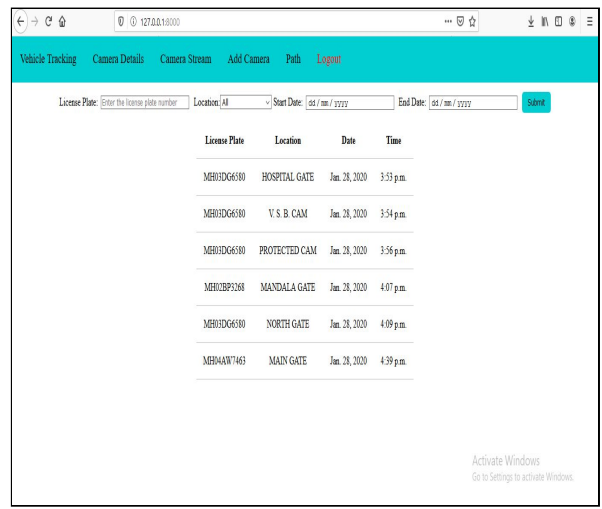

Fig 3b. Snapshots of GUI(2)

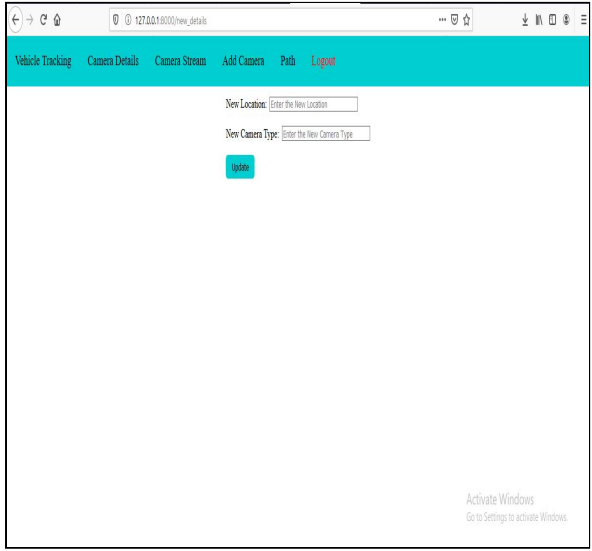

Fig 3c. Snapshots of GUI(3)

### B. Second Phase: License Plate Detector

In the second phase the frames captured by the camera are served as input to the detection module. In this module the system performs some basic preprocessing like resizing and passes the frame to the text detection model. The output of the model are bounding box coordinates, using these coordinates, the license plate is cropped and sent to the third phase. The architecture of the model this phase uses is mainly based on the Efficient and Accurate Scene Text Detector architecture(EAST)[8]. A few changes were made to the architecture to improve accuracy, as listed below. Instead of using VGGnet[9] as a feature extractor Resnet50[10] was used, also instead of using cross entropy as a loss function dice loss function was used, in place of Adam[11] optimizer AdamW[12] was used as an optimizer. The model was trained on the ICDAR 2015 and 2013 for 800 epochs. It achieves 0.802 F-score on the ICDAR 2015 test set. The model predicts a set of bounding boxes, which are then passed through non-max suppression, as mentioned in [8], to remove the less accurate boxes.

The model gives a set of boxes, out of which the less accurate ones are removed using non max suppression which use Intersection over Union values as its measure.

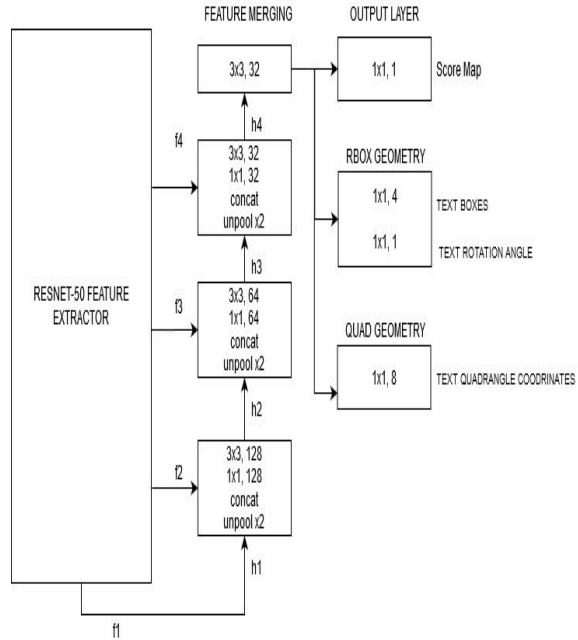

Fig 4. License plate detector model architecture

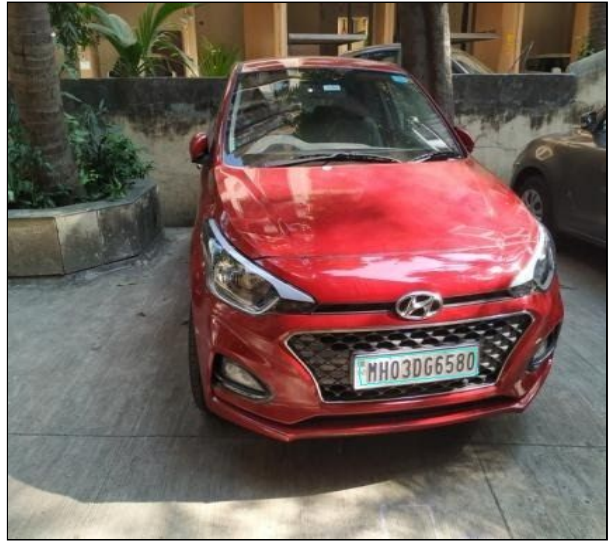

Fig 5. Output of Phase 2

### C. Third Phase: Character Recognition

In the third phase, the cropped license plate image is accepted as input, a set of preprocessing operations are performed on the input and the output is fed to the custom trained Tesseract OCR engine.

The preprocessing is done in order to make the image suitable for the Tesseract to process. Preprocessing operations performed are Gaussian Blurring, Histogram Equalisation, Binary Thresholding and Character

Segmentation. The final output is sent to the Tesseract Engine which is custom trained and also has been provided with a whitelisted character.

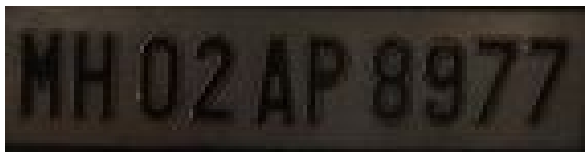

Fig 6a. Original Cropped plate

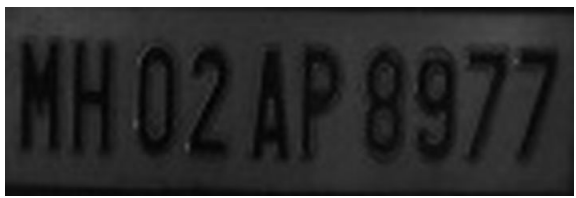

Fig 6b. Gaussian Blur Output

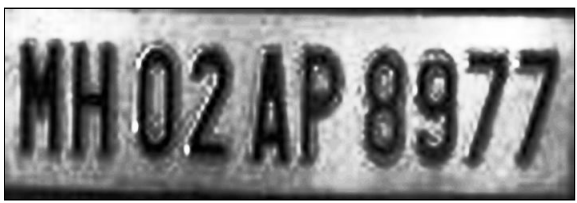

Fig 6c. Histogram Equalization output

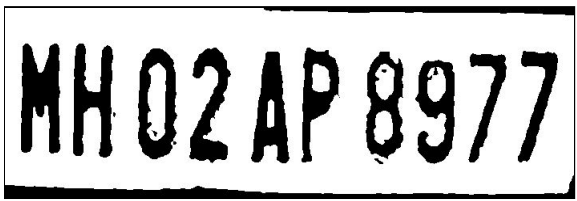

Fig 6d. Binary Threshold output

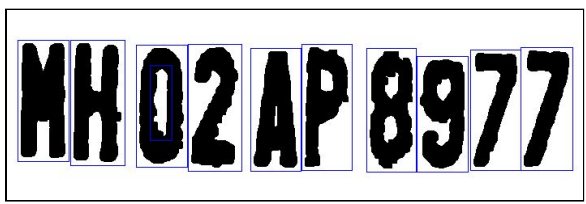

Fig 6e. Character Segmentation

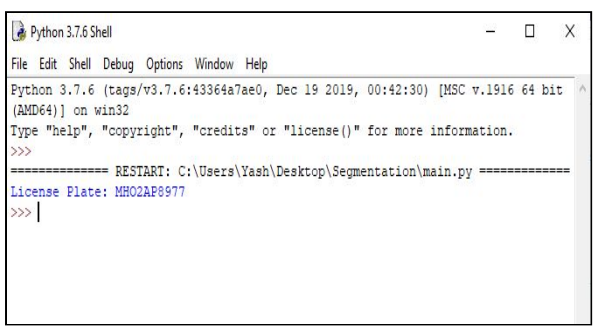

Fig 7. Tesseract Output

## VI. RESULTS AND ANALYSIS

The proposed system provides quite good results when it comes to detecting license plates, but when it comes to the recognition part the tesseract fails in a few cases even if the cropped image is clear. It is due to the fact that many font styles are used in making license plates and tesseract is not trained on all of them. The Detection module is capable of producing results at a speed of 33 fps on NVIDIA Geforce RTX 2080 ti, for all orientations of the license plate.

Since the detection module is based on a text detection model other scene text are also detected but that can be eliminated by applying some filters like pixel value or the size of the text area etc.

## VII. CONCLUSION AND FUTURE SCOPE

The system implements the solution to the problem of vehicle tracking using license plate detection and recognition, as license plate numbers are the sole unique identifiers of a vehicle. For the purpose of detection the system uses scene text detection method, as it is easy to implement and produces accurate results. For the task of license plate recognition the system makes use of Tesseract. It also provides a user customized website. The results are impressive but do not provide a high accuracy level.

The future scope for this system is high. As the concept of self driving cars is coming to reality, the need for VLPR and vehicle tracking increases. Further this system can be extended to be used for monitoring and charging fines to traffic rule violators and can also be used for finding suspicious/stolen vehicles. The system can also be integrated with the RTO database, due to which the relevant personnel can get the details of the owner and also verify if the license plate is forged or not.

## VIII. REFERENCES

[1] Prof. Pradnya Randive, Shruti Ahivale, Sonam Bansod, Sonal Mohite, Sneha Patil, "Automatic License Plate Recognition [Alpr]-a Review Paper",International Research Journal of Engineering and Technology (IRJET), Jan 2016.

[2] Rayson Laroca et al, "A Robust Real-Time Automatic License Plate Recognition Based on the YOLO Detector", arXiv:1802.09567v6 [cs.CV], 28 Apr 2018.